\documentclass[preprint,12pt,authoryear]{elsarticle}

\usepackage{amssymb}
\usepackage{amsmath}
\usepackage{caption}
\usepackage{hyperref}
\usepackage{graphicx}
\usepackage{latexsym}
\usepackage{times}
\usepackage{siunitx}
\usepackage[pagewise]{lineno}
\usepackage[T1]{fontenc}

\usepackage{color}

\usepackage{lineno}

\makeatletter
\def\ps@pprintTitle{%
    \let\@oddhead\@empty
    \let\@evenhead\@empty
    \def\@oddfoot{\reset@font\hfil}%
    \let\@evenfoot\@oddfoot
}
\makeatother

\begin{document}

\begin{frontmatter}


\title{Flow reconstruction in time-varying geometries using graph neural networks} 

\author[a]{Bogdan A. Danciu\corref{cor1}}
\author[a]{Vito A. Pagone}
\author[b]{Benjamin Böhm}
\author[b]{Marius Schmidt}
\author[a]{Christos E. Frouzakis}

\affiliation[a]{organization={CAPS Laboratory, Department of Mechanical and Process Engineering, ETH Zürich},
               city={Zürich},
               postcode={8092},
               country={Switzerland}}

\affiliation[b]{organization={Technical University of Darmstadt, Department of Mechanical Engineering, Reactive Flows and Diagnostics},
               addressline={Otto-Berndt-Str. 3},
               city={Darmstadt},
               postcode={64287},
               country={Germany}}

\cortext[cor1]{Corresponding author}

\begin{abstract}

The paper presents a Graph Attention Convolutional Network (GACN) for flow reconstruction from very sparse data in time-varying geometries. The model incorporates a feature propagation algorithm as a preprocessing step to handle extremely sparse inputs, leveraging information from neighboring nodes to initialize missing features. In addition, a binary indicator is introduced as a validity mask to distinguish between the original and propagated data points, enabling more effective learning from sparse inputs. Trained on a unique data set of Direct Numerical Simulations (DNS) of a motored engine at a technically relevant operating condition, the GACN shows robust performance across different resolutions and domain sizes and can effectively handle unstructured data and variable input sizes. The model is tested on previously unseen DNS data as well as on an experimental data set from Particle Image Velocimetry (PIV) measurements that were not considered during training. A comparative analysis shows that the GACN consistently outperforms both a conventional Convolutional Neural Network (CNN) and cubic interpolation methods on the DNS and PIV test sets by achieving lower reconstruction errors and better capturing fine-scale turbulent structures. In particular, the GACN effectively reconstructs flow fields from domains up to 14 times larger than those observed during training, with the performance advantage increasing for larger domains.

\end{abstract}

\begin{keyword}
Graph neural networks, Time-varying geometries, Internal combustion engines, Direct numerical simulations, Particle image velocimetry
\end{keyword}

\end{frontmatter}



\section{Introduction} \label{sec:introduction}

The reconstruction of spatial fields from sparse and limited data is a major challenge in the analysis, estimation and control of complex physical systems. In various fields such as atmospheric research~\citep{TelloAlonso2010, mishra2014}, autonomous aerial navigation~\citep{Achermann2019, Achermann2024} and fluid dynamics~\citep{fukami2019}, conventional linear methods such as linear stochastic estimation~\citep{adrian_moin_1988}, Delaunay triangulation~\citep{Saini2016} and proper orthogonal decomposition~\citep{thanh2004, Druault2005}
face challenges in accurately reconstructing extensive spatial patterns. This is especially true when dealing with very sparse data and problems such as system nonlinearity and boundary effects.
Neural Networks (NN) have emerged as a promising nonlinear alternative that has proven to be effective in efficiently reconstructing chaotic data from sparse information. Building on this foundation, Machine Learning (ML) has achieved remarkable success in generating flow fields using data from experimental observations and numerical simulations (e.g. \cite{kissas2020, yan2019, cai2019}).

Recent work has demonstrated the power of ML approaches in fluid dynamics applications. \cite{Morimoto2021} used a Convolutional Neural Network (CNN) to analyze artificial Particle Image Velocimetry (PIV) data, and propose a novel approach for reconstructing flow fields from snapshots containing regions with missing data. \cite{Kochkov2021} employed an end-to-end CNN-based model to improve approximations inside Computational Fluid Dynamics (CFD) domains for modeling two-dimensional (2D) turbulent flows. \cite{manickathan22} proposed CNN as an alternative to the image cross-correlation methods commonly used in processing PIV data to reconstruct the fluid velocity field, and it was shown to outperform conventional methods in terms of robustness to data noise and providing reconstructed velocity fields with significantly higher spatial resolution. Super-Resolution (SR) methods have also been used to reconstruct high-resolution flow fields from low-resolution data. \cite{fukami2019} utilized a CNN and a hybrid downsampled skip-connection/multi-scale (DSC/MS) CNN model, for the SR analysis of turbulent flow fields from very coarse data. \cite{kim2021} used the cycle-consistent generative adversarial network (CycleGAN) to reconstruct flow fields from low-resolution DNS and Large Eddy Simulation (LES) data. \cite{Bode2021, Bode2023} developed a physics-based super-resolution generative adversarial network (PIESRGAN) for subfilter-scale turbulence reconstruction that uses a loss function based on the continuity equation residue. Even though only homogeneous isotropic data was used to train the model, it was able to make better predictions of scalar mixing in a reacting jet.

The aforementioned models rely on convolutional layers and can therefore be applied to unstructured data only to a limited extent. Since conventional ML methods require a feature matrix with a specific size and order of input samples, they cannot be readily applied to unstructured data. However, flow field data can be highly unstructured due to irregular meshes in curved or complex geometries. Several approaches have been developed to address the limitations of CNN when handling unstructured data in fluid dynamics. \cite{Heaney2024} proposed using space-filling curves to transform multi-dimensional solutions on unstructured meshes into a one-dimensional (1D) representation, allowing for the application of 1D convolutional layers. \cite{Xu2021} developed an unstructured CNN that aggregates and exploits features from neighboring nodes through a weight function, enabling convolutions on irregular grids. \cite{Kashefi2021, Kashefi2022} explored point cloud deep learning frameworks where CFD grid vertices are treated as point clouds and used as inputs to neural networks. Graph Convolutional Networks (GCN) have also emerged as a powerful tool for unstructured data. \cite{He_2022} introduced the flow completion network (FCN) that employs a GCN to deduce fluid dynamics from sparse data sets. \cite{duthe2023graph} also used a GCN to predict the flow field and far-field boundary conditions based on the pressure distribution at the surface of airfoils. 

Despite these advancements, several challenges remain in applying ML techniques to fluid dynamics problems. One significant limitation is the difficulty in handling extremely sparse data, where traditional interpolation methods often fail to capture complex flow features. Additionally, time-varying geometries pose a unique challenge, as most current ML models are designed for static configurations and struggle to adapt to dynamic changes in the flow domain. Another critical issue is the generalization of ML models across different types of data sets, from high-fidelity numerical simulations to experimental measurements. The discrepancies in data quality, resolution, and underlying physics between these sources can lead to poor accuracy when models trained on one type of data are applied to another.

In this paper, we introduce a Graph Attention Convolutional Network (GACN) trained on a unique data set of three-dimensional (3D) Direct Numerical Simulations (DNS) modeling the compression-expansion stroke of a single-cylinder Internal Combustion Engine (ICE) under practically relevant operating conditions. The DNS data presents specific challenges for conventional ML applications due to its unstructured grid that dynamically changes during compression, leading to variations in resolution. GACN effectively addresses both aspects: the unstructured nature of the data is captured by the position of the graph nodes, while the changing resolution is accounted for by the distance features of the edges between nodes. Furthermore, we present a method to handle extremely sparse data, which performs remarkably well even when 99\% of the data is missing. 
This is achieved by incorporating a Feature Propagation (FP) algorithm as a preprocessing step and adding a Binary Indicator (BI) as an extra feature. The FP algorithm initializes absent features with values that are both reasonable and physically consistent, leveraging data from neighboring nodes. The BI serves as a validity mask, providing crucial information to the network about which data points are original and which are propagated, enabling a more effective learning from sparse inputs without the limitations of using default values for missing data. Notably, we demonstrate the robustness and versatility of the approach by successfully applying the model trained on DNS data to experimental PIV measurements, achieving promising results despite the inherent differences between numerical and experimental data sources.

The rest of the paper is organized as follows. Section~\ref{sec:datasets} introduces the DNS and PIV data sets used for training and evaluating the model. In Sec.~\ref{sec:methodology}, we provide a detailed description of the GACN model, as well as the alternative models used for comparison, specifically a CNN and classical cubic interpolation. Section~\ref{sec:results} presents a comprehensive analysis of the model predictive accuracy using both numerical data derived from DNS simulations and experimental data from PIV measurements which was not used during training. We compare the performance of the GACN model to a CNN architecture for which data was interpolated onto a uniform mesh and classical cubic interpolation. The main conclusions and future research directions are outlined in Sec.~\ref{sec:conclusions}. 

\section{Data Sets \label{sec:datasets}}

\subsection{DNS data \label{subsec:dns_data}}

\begin{figure}[h!]
\centering
\includegraphics[width=0.6\linewidth]{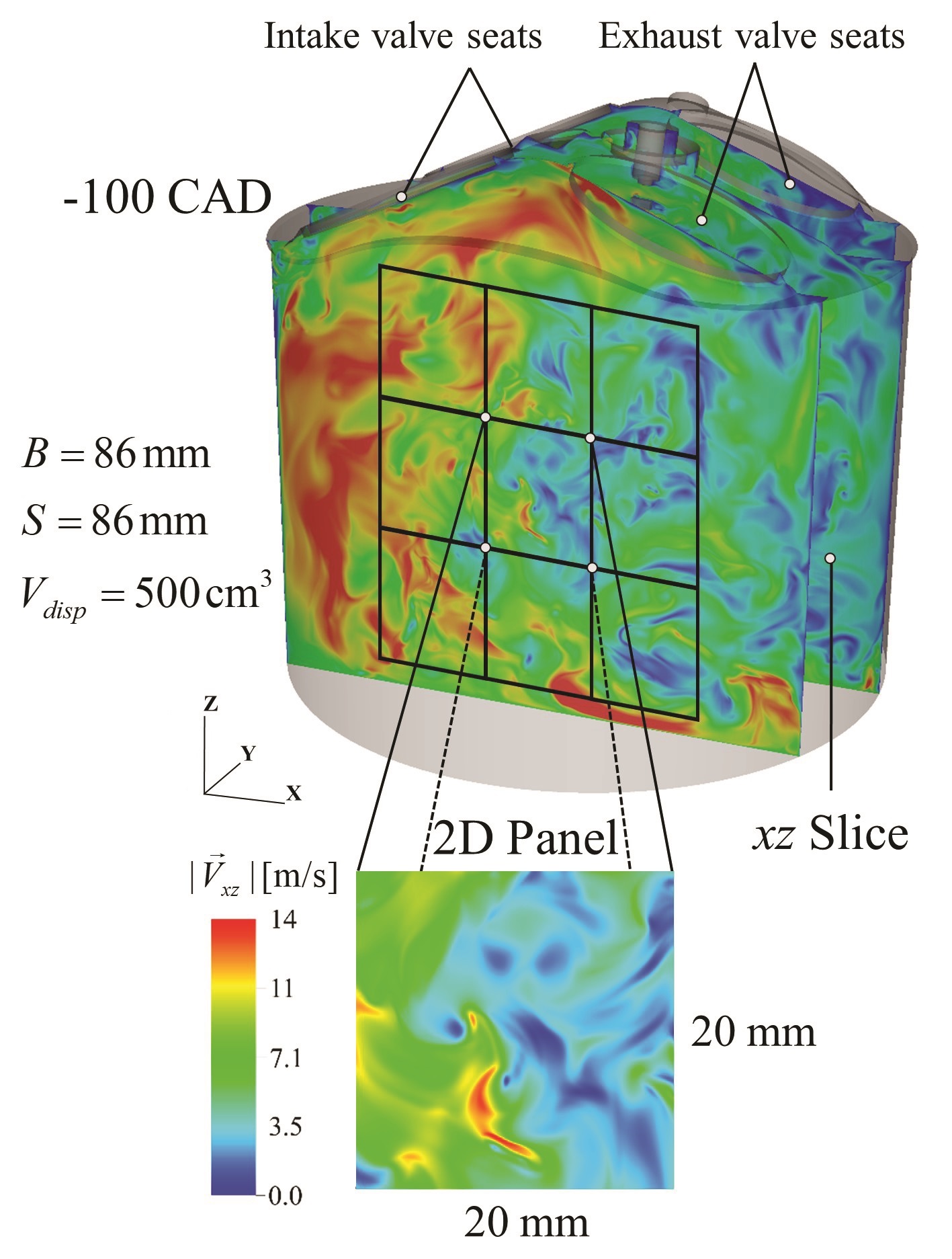}
\caption{\footnotesize Engine geometry and illustration of data generation showing the velocity magnitude distribution on three $xz$ slices.}
\label{fig:data_slice}
\end{figure}

The DNS data set consists of a multi-cycle simulation of a motored laboratory-scale engine at practically relevant conditions (1500 rpm and full load 0.95 bar intake pressure) \citep{Danciu2023, Danciu2024}.
The single-cylinder, optically-accessible engine studied at TU Darmstadt has a four-valve pent-roof cylinder head and an intake duct that promotes the formation of a tumbling flow. The cylinder of the square engine has a bore  of $B=86$~mm. The computational domain is shown in Fig.~\ref{fig:data_slice}.

\addvspace{20pt}

Detailed information on the experimental setup can be found in \cite{baum2014}. The DNS were performed with the spectral element solver NekRS \citep{nekrs} extended at ETHZ for moving geometries. In order to ensure optimal resolution over the entire cycle, four conformal hexahedral grids were constructed using Coreform Cubit (version 2022.11) with the number of spectral elements ranging from $E=4.8$ to $E=9.3$ million to accommodate the increase in the Reynolds number and the decrease in turbulence scales during compression. Using a polynomial order of $N=7$ the meshes yielded between 1.5 to 3.2 billion individual grid points. The average resolution was \SI{30}{\micro\metre} in the bulk, with the first grid point being \SI{3.75}{\micro\metre} away from the walls. The initial conditions were derived from precursor LES validated against experimental data \citep{Giannakopoulos2022}, providing a realistic starting point for the DNS.
A total of 12 compression-expansion cycles were computed on JUWELS Booster GPU nodes at the J\"{u}lich Supercomputing Centre using 70 GPU nodes, with each node equipped with four NIVIDIA A100 GPUs and requiring 0.13 million core-hours (MCore-h) per cycle \citep{Danciu2024}.

Using a complex DNS data set with multiple temporally-evolving unstructured grids poses significant difficulties for conventional ML models based on convolutional layers. The data would need to be interpolated on a structured grid and the variable input would need to be padded in order to be efficiently used as input to a CNN. On the other hand, the GACN can directly incorporate the unstructured grid points by including their coordinates as node features. The movement of the grid and the change of resolution can be captured by the graph edges, which include the distance between the nodes as features.

\subsection{PIV data\label{subsec:piv_data}}

High-speed PIV was used to measure the two in-plane velocity components of the flow field in the central tumble plane of the engine. The beam of a dual-cavity Nd:YAG laser (IS4II-DE Edgewave) operating at 4.8 kHz, frequency doubled to \SI{532}{\nano\metre}, was passed through a combination of a telescope and a negative cylindrical lens to form a light sheet (0.8~mm thick at 13.5\% of maximum intensity) introduced through the piston window, coaxial to the cylinder centerline via the Bowditch extension and piston mirror. The crank-angle dependent time interval $\Delta t$ between pulses was set between 3--51~$\mu s$ to achieve optimal particle displacement and minimize out-of-plane losses. DOWSIL 510 (Dow Corning) silicone oil was atomized by a fluid seeder (AGF 10.0, Palas) with an average particle size of \SI{0.5}{\micro\metre} and introduced into the intake system as tracer particles. A high-speed CMOS camera (Phantom v1610) with a Sigma lens (105~mm F2.8 Macro, f/11) was used to record the Mie scattering of the tracer particles perpendicular to the light sheet through the quartz glass cylinder.

DaVis 8.4 (LaVision) was used to calculate the flow fields using cross-corre\-la\-tion with multiple pass iterations with decreasing window size (twice: $64\times64$ pixels, 50\% overlap; twice: $32\times32$ pixels, 75\% overlap), a peak ratio threshold of 1.3, and a universal outlier median filter to remove false vectors. Possible sources of error and statistical uncertainties are discussed in detail in \cite{haussmann2020}. The  instantaneous velocity magnitudes uncertainty is estimated to be about 5\%.

In our study we only use the PIV data for testing purposes, as the GACN model never sees it during training. In this way, we can test the generalization capabilities of the network trained on numerical data to make predictions on experimental measurements.

\subsection{Training, validation and test data sets\label{subsec:training_data}}

Two-dimensional slices along the $xz$ tumble plane were extracted from the 3D DNS data. A total of 19 slices, spaced 4~mm apart, were selected along the $y$-direction. To maintain reasonable computational cost, depending on the piston position, up to 9 panels were selected on each slice (Fig.~\ref{fig:data_slice}). 
Due to the dynamic piston movement, the number of grid points in each panel varies between 130,000 and 470,000 during mid-compression. Panels were sampled every 5 crank-angle degrees (CAD) from -120 CAD (before top dead center) up to -60 CAD. 
The bulk Reynolds number, $Re=B\bar{V_p}/\nu$, calculated using the mean piston speed $\bar{V_p}$, bore $B$, and the kinematic viscosity $\nu$ of the gas, ranges between 35,000 and 120,000. Of the 12 simulated cycles, 10 were allocated for training and validation, while 2 were reserved for testing. In total, 9,754 panels were created for training and validation, and 2,237 panels for testing. Additionally, 840 $xz$ slices covering the whole vertical extent were selected exclusively for testing purposes. To evaluate model performance on experimental data across different resolutions and domain sizes, panels and complete $xz$ slices were selected from the central tumble plane of the PIV data. A total of 612 panels and 204  slices from 17 cycles were sampled every 5 CAD from -120 CAD to -60 CAD. Table~\ref{tab:train_test} summarizes the distribution of data into training, validation, and test sets.

\begin{table}\footnotesize
\caption{Distribution of data across training, validation, and test sets. DNS panels were split for model development, while DNS slices, PIV panels, and PIV slices were exclusively used for testing.}
\centerline{\begin{tabular}{c c c c}
\hline 
Data          & Train & Validation & Test   \\
\hline
DNS panels    & 7803  & 1950     & 2237   \\
DNS slices    &  -    &  -       & 840   \\
PIV panels    &  -    &  -       & 612   \\
PIV slices    &  -    &  -       & 204    \\
\hline 
\end{tabular}}
\label{tab:train_test}
\end{table}

\section{Methodology \label{sec:methodology}}

\subsection{The GACN model}

\subsubsection{Input and label composition\label{subsec:gacn_input_label}}

The input and label for the GACN model consist of graphs, where each node corresponds to a grid point as described in Sec.~\ref{subsec:training_data}. The nodes carry five features: the two velocity components ($u_x$ and $u_z$), the binary indicator BI and the $x,z$ coordinates. The distance between the nodes is represented by the graph edges that carry the corresponding distance feature.
To create the sparse input for testing, the features for 99\% of the nodes are randomly set to zero. The BI flags nodes that contain non-zero features. Previous work \citep{kohler2014, uhrig2017, Achermann2024} has shown the value of providing binary input masks for CNN that handle sparse input data, and we have extended the concept for GCN. The network output retains the graph structure, and focuses on predicting the two velocity components for each node. This effectively reconstructs the velocity field while preserving the spatial information encoded in the original graph structure.

\subsubsection{GACN architecture\label{subsec:nn_architecture}}

\begin{figure*}[h!]
\centering
\includegraphics[width=\linewidth]{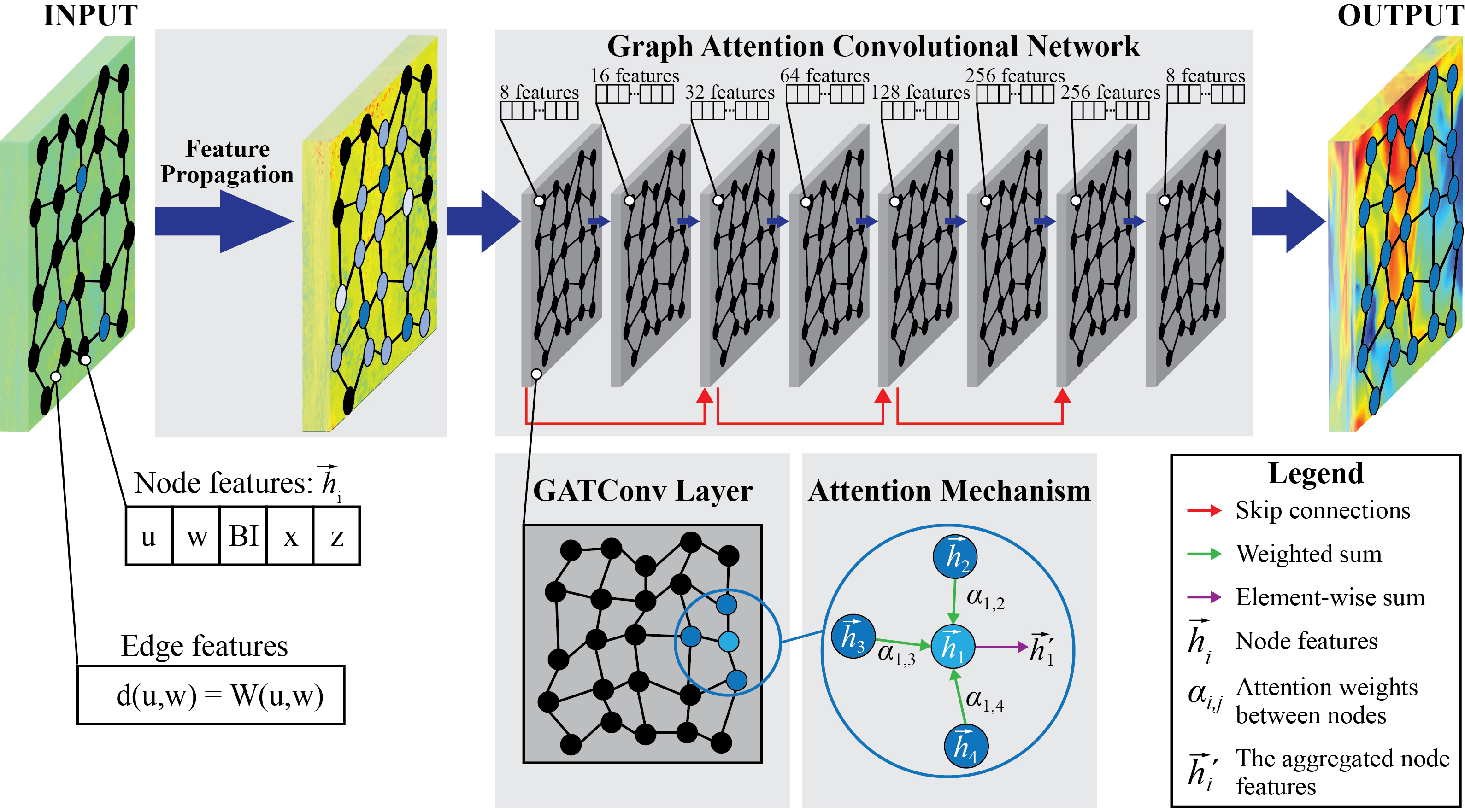}
\caption{\footnotesize Schematic of the GACN model architecture and its internal structure.}
\label{fig:gacn_architecure}
\end{figure*}

The GACN architecture is shown in Fig.~\ref{fig:gacn_architecure}. We use the FP algorithm proposed by \cite{rossi2022on} as a preprocessing step before feeding the input to the network. The algorithm is an efficient iterative process for reconstructing missing node features in graphs, and operates on the principle of Dirichlet energy minimization to diffuse known features throughout the entire graph and effectively reconstruct missing features even when a large fraction of the features is not available. 

Inspired by the work of \cite{velickovic}, the model architecture contains eight layers of PyTorch Geometric Graph Attention Convolution (GATConv) modules, which incrementally expand the feature dimensions from 8 to 256. 
The attention mechanism in GATConv operates by computing attention coefficients for each node with respect to its neighbors. These coefficients are derived from a learnable attention function that takes as input the features of both the node and its neighbor. The attention weights are then obtained by applying a softmax function to these coefficients, ensuring they sum to 1 for each node. The aggregated feature for a node is computed as a weighted sum of its neighbors' features, where the weights are the computed attention weights. Formally, if we denote the feature of node $i$ as $\bar{h}_i$, and the set of its neighbors as $\mathcal{N}(i)$, the aggregated feature $\bar{h}'_i$ is calculated as:
\begin{equation}
\bar{h}'_i = \sigma\left(\sum_{j \in \mathcal{N}(i)} \alpha_{ij} \mathbf{W} \bar{h}_j\right)
\end{equation}
where $\alpha_{ij}$ is the attention weight between nodes $i$ and $j$, $\mathbf{W}$ is a learnable weight matrix, and $\sigma$ is a non-linear activation function. This mechanism allows the network to adaptively focus on the most relevant neighbors for each node, potentially capturing complex, non-local dependencies in the graph structure. The attention mechanism introduces flexibility and adaptivity in learning the underlying patterns and structures within the graph, especially in complex data cases like fluid dynamics. Additionally, the model strategically implements skip connections \citep{pmlr-v139-xu21k} that accelerate training and help improve performance with increased depth, leveraging both local and global graph information. In addition, the network integrates a Laplacian diffusion process with a learnable parameter $\alpha$, diffusing node features in accordance with the graph Laplacian. This approach allows for adaptive feature propagation, enhancing the network ability to capture and represent the underlying graph structure more effectively, and embed a deeper comprehension of graph topology into the feature learning process.

The resulting GACN model contains 148,881 trainable parameters, balancing model complexity with computational efficiency.

\subsubsection{Loss function} \label{subsec:gacn_loss_function}

For the GACN model, we employ as the primary optimization criterion the Mean Squared Error (MSE) ${L}_{MSE}$, i.e. the $L_2$ norm of the error to ensures that the network predictions closely match the actual velocities:
\begin{equation}
\mathcal{L}_{MSE} = \frac{1}{N}\sum_{i=1}^{N}(u_i - \hat{u}_i)^2. \label{eq:mse_loss}
\end{equation}
Here, $u_i$ and $\hat{u}_i$ denote the actual and the predicted velocity at graph node $i$,  respectively. 

\subsection{CNN model used for comparison \label{subsec:cnn_architecture}}

\subsubsection{Input and label composition} \label{subsec:cnn_input_label}

The input to the CNN model consists of three features: the two velocity components ($u_x$ and $u_z$) and a binary mask that serves a purpose similar to the BI in the GACN model, marking the points that contain valid data or are zero-valued. To prepare the data for the CNN, we interpolated the DNS panels to a uniform $512\times512$ grid, resulting in 262,144 points. This number is chosen to match the average number of nodes used as inputs to the GACN model (about 250,000 nodes per panel.) To simulate sparse data conditions and maintain consistency with the GACN input, we randomly set 99\% of the input velocity values to zero, retaining only 1\% of the original velocity data. The binary mask is then used to specify these non-zero velocity locations. This approach ensures a consistent resolution and sparsity level in both architectures and preserves the small-scale velocity features of the input data. The interpolation process maintains the spatial structure of the flow field while adapting it to the required input format for the CNN. The labels for the CNN model are, similar to the GACN, the two predicted velocity components.

\subsubsection{CNN architecture}

\begin{figure*}[h!]
\centering
\includegraphics[width=\linewidth]{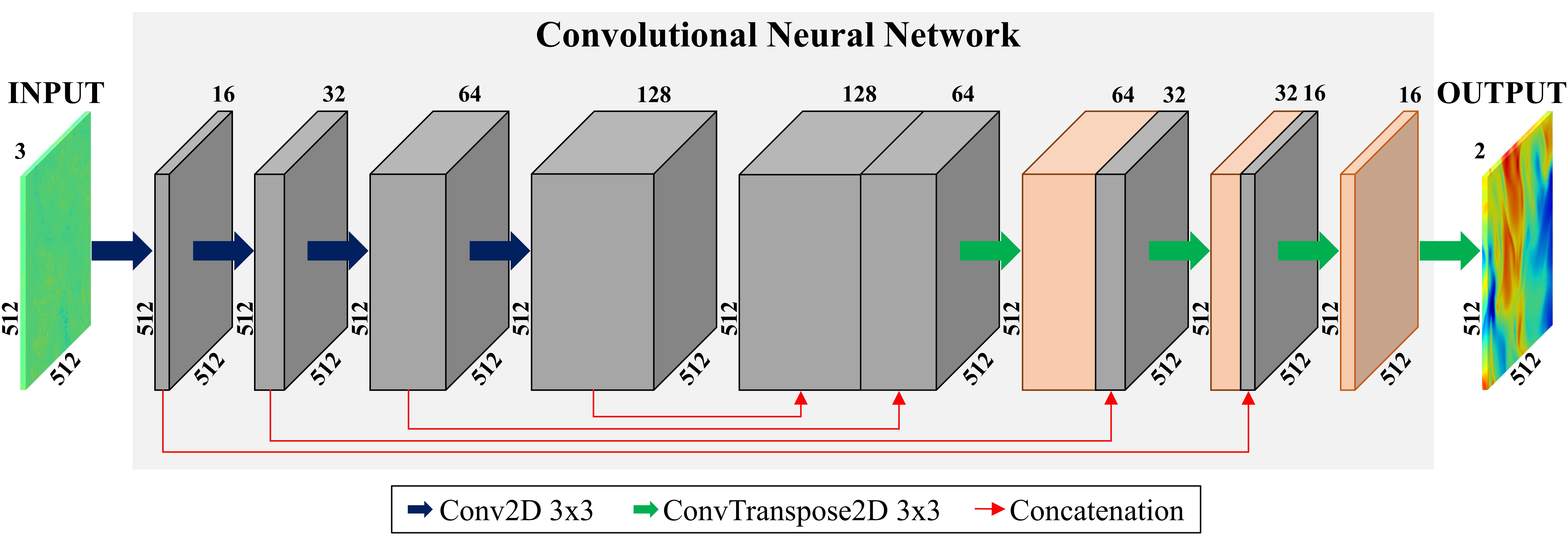}
\caption{\footnotesize The architecture of the CNN model used for comparison.}
\label{fig:cnn_architecture}
\end{figure*}

The CNN architecture consists of eight layers evenly distributed between the encoder and decoder parts (Fig.~\ref{fig:cnn_architecture}). The network input comprises three channels: two for the velocity components and one for the binary mask.
We opted against a U-shaped network architecture, as initial experiments have shown that such designs tend to produce more pronounced checkerboard artifacts in the reconstructed flow fields. Instead, the architecture maintains the spatial dimensions throughout all layers, preserving the original size of the input. The four-layer encoder section of the CNN gradually increases the channel depth from 4 to 128, allowing the network to capture complex hierarchical patterns in the data. Each of these layers consists of a convolutional layer followed by a ReLU activation function that ensures effective feature extraction. The decoder, which also consists of four layers, methodically reduces the channel depth back to two in order to adapt to the original velocity channels while maintaining spatial dimensions. This is achieved by transposed convolutional layers. To preserve important spatial details that might otherwise be lost during encoding, and to improve the network ability to accurately reconstruct patterns, we implement skip connections between the corresponding layers in the encoder and decoder. These connections allow the model to utilize both high-level abstract features and low-level detailed information during the reconstruction process.

The resulting CNN model contains 243,881 trainable parameters. This number is higher than that of the GACN model (148,881 parameters), but was intentionally kept comparable in size to ensure a fair comparison between the two architectures. 
This similarity in model size allows us to evaluate the differences in performance primarily in terms of architectural approaches rather than differences in model complexity.

\subsubsection{Loss function} \label{subsec:cnn_loss_function}

For the CNN, the loss functions for training consists of two components. The MSE loss ($\mathcal{L}_{MSE}$) (Eq.~\ref{eq:mse_loss}) is combined with the Total Variation (TV) loss $\mathcal{L}_{TV}$~\citep{Chambolle2009} defined as the sum of the absolute differences between neighboring pixel values,
\begin{equation}
\mathcal{L}_{TV} = \frac{2}{N}\sum_{i=1}^{N} \left( \left|u_{i+1, j} - u_{i, j} \right| + \left|u_{i, j+1} - u_{i, j} \right| \right),
\end{equation}
where $u_{i,j}$ represents the predicted velocity value at pixel $(i,j)$, and $N$ is the total number of pixels. 
This ensure that the predictions do not exhibit abrupt spatial changes, which is unlikely in real fluid flows. By penalizing high-frequency fluctuations, the $\mathcal{L}_{TV}$ term results in more coherent and realistic flow fields and avoids pixelated results. 
The incorporation of the TV loss not only improved the visual quality of the predictions but also enhanced the overall performance of the network. We observed that models trained with this loss demonstrated better generalization capabilities and produced more accurate reconstructions.

The complete loss function for the CNN is thus formulated as a weighted sum of the aforementioned losses:
\begin{equation}
\mathcal{L}_{total} = \mathcal{L}_{MSE} + \alpha \mathcal{L}_{TV},
\end{equation}
where the hyperparameter $\alpha$ controls the relative importance of the TV loss by balancing its contribution to the total loss. 

\subsection{Implementation details\label{subsec:implementation_details}}

The GACN and CNN models were implemented with Python, using the PyTorch~\citep{PyTorch} and PyTorch Geometric~\citep{fey2019fast} libraries. PyTorch Geometric, a specialized extension of PyTorch, was crucial for the efficient handling of graph data structures and facilitated the training and evaluation of the GACN model. We trained the GACN for 100 epochs and the CNN for 150 epochs. The training processes started from scratch by randomly initializing the weights of the networks using the Glorot (Xavier) initializer~\citep{Glorot2010}, with the biases initialized to zero. Both models were trained end-to-end using the Adam optimizer~\citep{Kingma2014} with an initial learning rate of $10^{-4}$. To optimize the learning process, we employed an adaptive learning rate strategy that reduced the rate when validation performance plateaued. This approach, commonly known as learning rate decay, helps to fine-tune the convergence of the model.
The training was performed on a single node of the JUWELS Booster GPU cluster at the Jülich Supercomputing Centre, equipped with 2 AMD EPYC Rome CPUs (48 cores) and 4 NVIDIA A100 GPUs, each with 40~GB of memory. The training process required 48 and 50 node hours for the CNN and GACN models, respectively.

\subsection{Performance metrics}

To evaluate the accuracy of the model in reconstructing the flow fields, we used two complementary metrics: Mean Absolute Error (MAE) and Root Mean Squared Error (RMSE).
MAE measures the average magnitude of errors in a set of predictions and is defined as,
\begin{equation}
\text{MAE} = \frac{1}{N} \sum_{i=1}^{N} |u_i - \hat{u}_i|.
\end{equation}
RMSE provides a quadratic scoring rule that measures the average error magnitude. It is particularly useful in our case as it assigns higher weight to large errors, which is relevant for capturing significant deviations in flow structures. It is defined as
\begin{equation}
\text{RMSE} = \sqrt{\frac{1}{N} \sum_{i=1}^{N} (u_i - \hat{u}_i)^2}.
\end{equation}
The  individual metrics are averaged across all test cases to provide an overall assessment of the model performance.

\section{Results and discussion\label{sec:results}}

\subsection{Ablation study\label{subsec:ablation_study}}

We conducted an ablation study to understand the impact of different components in the GACN model focusing on the effects of the FP algorithm, the BI, and the graph attention mechanism. Table~\ref{tab:ablation_fp_bi} shows the model performance with different 
combinations of the FP algorithm and BI on the DNS panel test set consisting of 2237 samples (Table~\ref{tab:train_test}). The results demonstrate that both the FP algorithm and the BI contribute to improved model performance, with the combination of both yielding the best results.

\begin{table}[h]
\caption{Ablation study results for FP and BI}
\label{tab:ablation_fp_bi}
\centering
\begin{tabular}{lccc}
\hline
Model Configuration & MAE [\SI[per-mode=symbol]{}{\meter\per\second}] & RMSE [\SI[per-mode=symbol]{}{\meter\per\second}]\\
\hline
GACN without FP and BI & 0.242 & 0.382 \\
GACN With FP & 0.194 & 0.314 \\
GACN With FP and BI & 0.156 & 0.247 \\
\hline
\end{tabular}
\end{table}

To further evaluate the effectiveness of the approach, we investigated the influence of different GNN architectures. We compared the graph attention layers with two alternatives: GraphSAGE layers \citep{Hamilton2017} and conventional graph convolutional layers \citep{Kipf2017}. This comparative analysis aims to isolate the effects of network architecture while keeping other model parameters constant. Table~\ref{tab:ablation_gnn} summarizes the results of this architectural comparison on the DNS panel test set and provides insights into the relative performance of each approach in flow field reconstruction. The GACN model outperforms both GraphSAGE and the more standard GCN in our task, indicating that the attention mechanism is particularly effective in capturing the relevant features for flow reconstruction.

\begin{table}[h]
\caption{Comparison of different graph neural network architectures}
\label{tab:ablation_gnn}
\centering
\begin{tabular}{lccc}
\hline
Graph Neural Network & MAE [\SI[per-mode=symbol]{}{\meter\per\second}] & RMSE [\SI[per-mode=symbol]{}{\meter\per\second}] \\
\hline
GACN & 0.156 & 0.247 \\
GraphSAGE & 0.201 & 0.334 \\
GCN & 0.287 & 0.429 \\
\hline
\end{tabular}
\end{table}

The ablation study highlights the importance of the FP algorithm and BI to significantly enhance the model ability to handle sparse input data, while the graph attention mechanism proves superior in learning the complex relationships within the flow field.

\subsection{Reconstruction of DNS data\label{subsec:dns_results}} 

\begin{figure*}[h!]
\centering
\includegraphics[width=\linewidth]{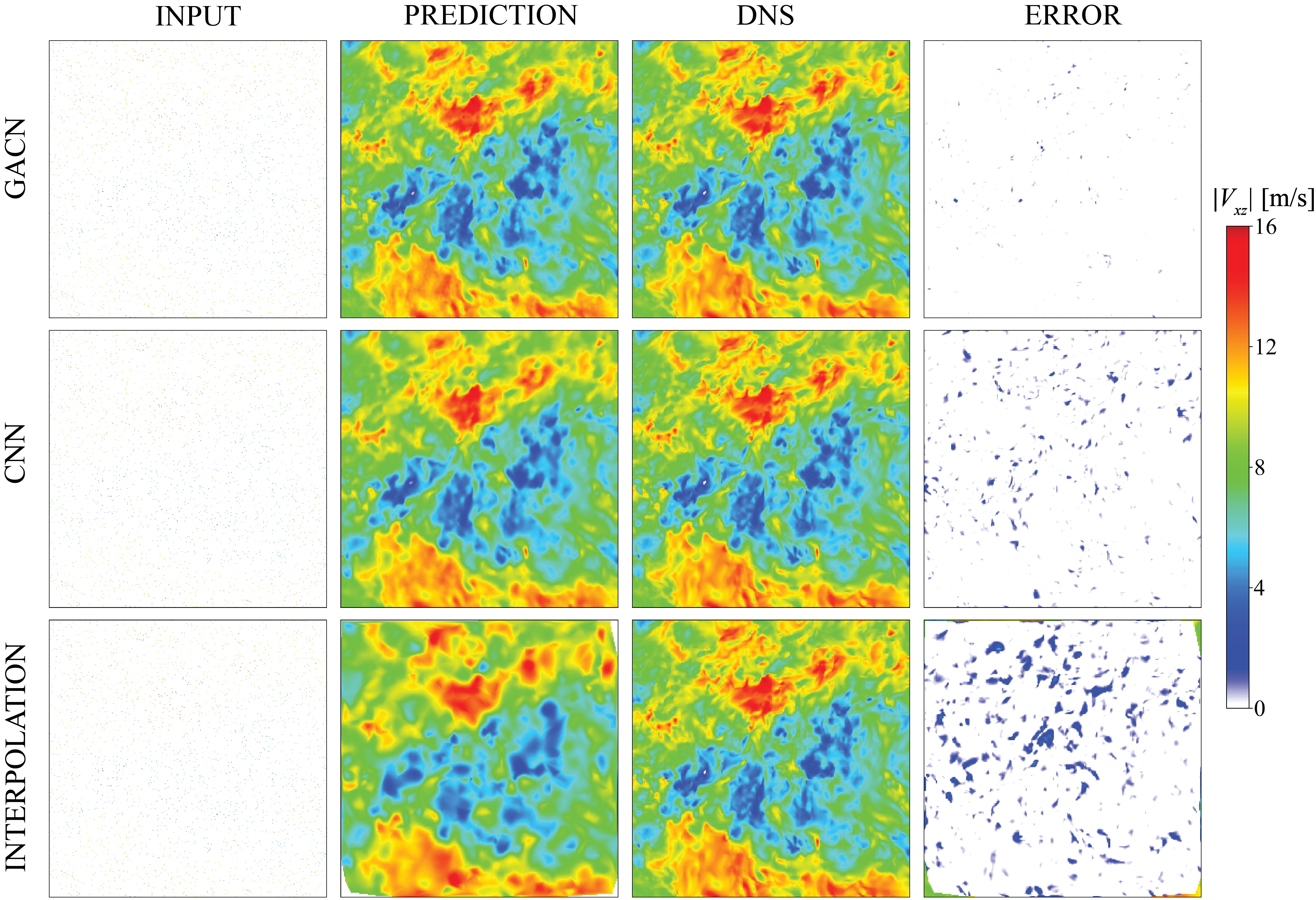}
\caption{\footnotesize Results for one panel at -60 CAD from the DNS panel test set. Rows from top to bottom show: GACN, CNN, and interpolation results. Columns from left to right present: input with 1\% retained information, model prediction, DNS ground truth, and absolute error between prediction and ground truth.}
\label{fig:dns_panel}
\end{figure*}

Figure \ref{fig:dns_panel} illustrates the reconstruction capabilities of the GACN model compared to two baseline methods: standard cubic interpolation, a common benchmark for modern ML-based flow reconstruction methods \citep{fukami2021voronoi, xu2023tranformer, Kazemi2023}, and the more competitive CNN model presented in Sec.~\ref{subsec:cnn_architecture}. The input for this comparison comes from a test panel with 99\% of the node velocity features set to zero. This sparse input is fed directly into the CNN and interpolation methods. For the GACN, the input is preprocessed by the FP algorithm to initialize the flow field before it enters the network. The results show that the ML-based methods significantly outperform cubic interpolation in flow reconstruction, with the GACN achieving better results overall. While the CNN has difficulty capturing the smallest turbulent scales and the finest features, the GACN shows a distinct ability to reconstruct these complex flow features. Cubic interpolation, although capable of capturing larger-scale features, it cannot accurately reconstruct smaller turbulent structures. In the last column of Fig.~\ref{fig:dns_panel}, the differences in accuracy are further highlighted by comparisons of the mean absolute error between the ground truth and predictions for each method.

\begin{figure*}[h!]
\centering
\includegraphics[width=\linewidth]{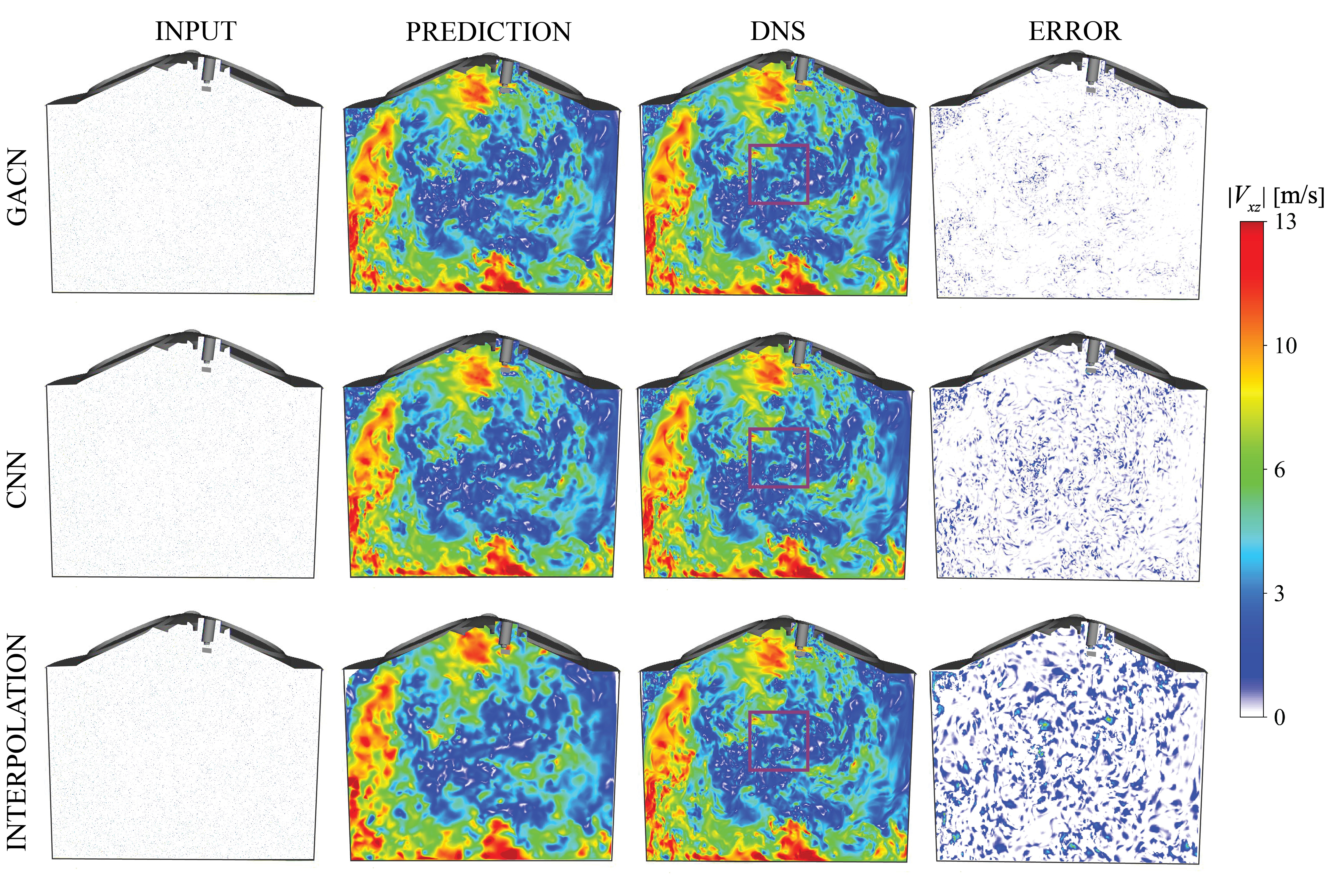}
\caption{\footnotesize Results for one slice at -90 CAD from the DNS slice test set. Rows from top to bottom show: GACN, CNN, and interpolation results. Columns from left to right present: input with 1\% retained information, model prediction, DNS ground truth, and absolute error between prediction and ground truth. The dark grey outline highlights the engine geometry. The purple square in the DNS column illustrates the size of a training panel and is not a velocity feature.}  \label{fig:dns_slice}
\end{figure*}

The reconstruction capabilities of the GACN model was also compared using as input an $xz$ slice that includes the cylinder head and the spark plug, the dark grey region from the slices in Fig.~\ref{fig:dns_slice}. The area that is approximately 14 times larger than the training data and contains about 3.6 million unique grid points, was processed with 99\% of the velocity features set to zero (Fig.~\ref{fig:dns_slice}.) The GACN processed the entire slice without data pre-processing.
For the CNN model, we interpolated the $xz$ slice to a uniform $2048\times2048$ grid, which is 16 times larger than the grid used in training, to meet the fixed input size requirements and obtain a similar resolution as for the GACN model. Despite the considerable increase in domain size, both networks accurately reconstruct the complex flow, with the GACN performing better. The cubic interpolation method exhibits larger errors in this extended domain.
The last column of Fig.~\ref{fig:dns_slice} shows the mean absolute error (MAE) between the ground truth and predictions.

\begin{figure}[h!]
\centering
\includegraphics[width=0.7\linewidth]{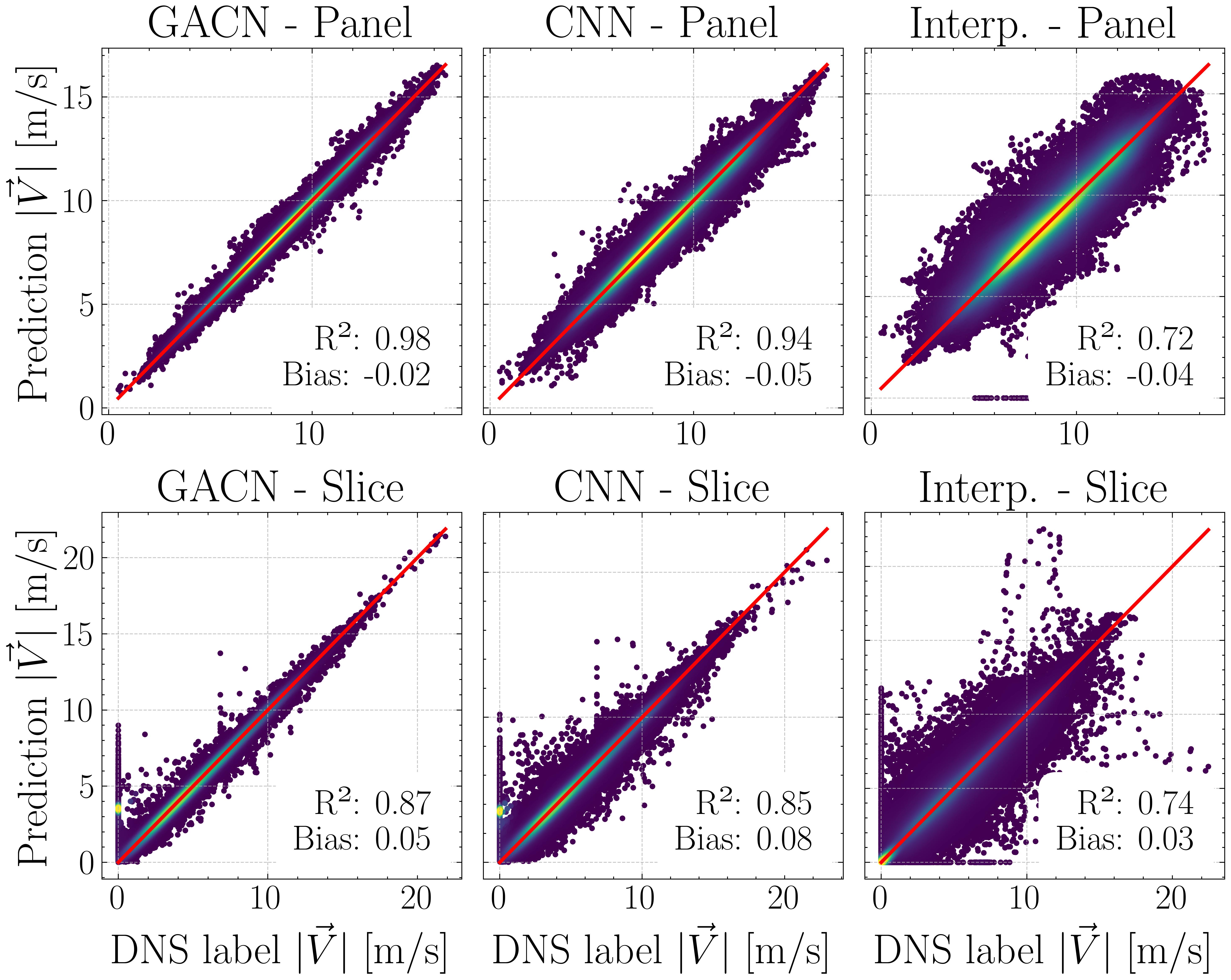}
\caption{\footnotesize Density scatter plots comparing the velocity magnitude label and prediction for the GACN (left), CNN (middle) and cubic interpolation (right) for the panels presented in Fig.~\ref{fig:dns_panel} (top) and the slices in Fig.~\ref{fig:dns_slice} (bottom).}
\label{fig:dns_scatter}
\end{figure}

The scatter plots of the predictions from Figs.~\ref{fig:dns_panel} and \ref{fig:dns_slice} depicted in Fig.~\ref{fig:dns_scatter} provide a direct comparison  and emphasize the superior predictive capabilities of the GACN model. For both the analyzed panel and the larger slice, the GACN model shows a tighter clustering of points around the diagonal. Its higher $R^2$ scores compared to the CNN and interpolation methods for the analyzed panel and slice show that a larger proportion of the variance of the true velocities is captured by the model.

\begin{figure}[h!]
\centering
\includegraphics[width=0.6\linewidth]{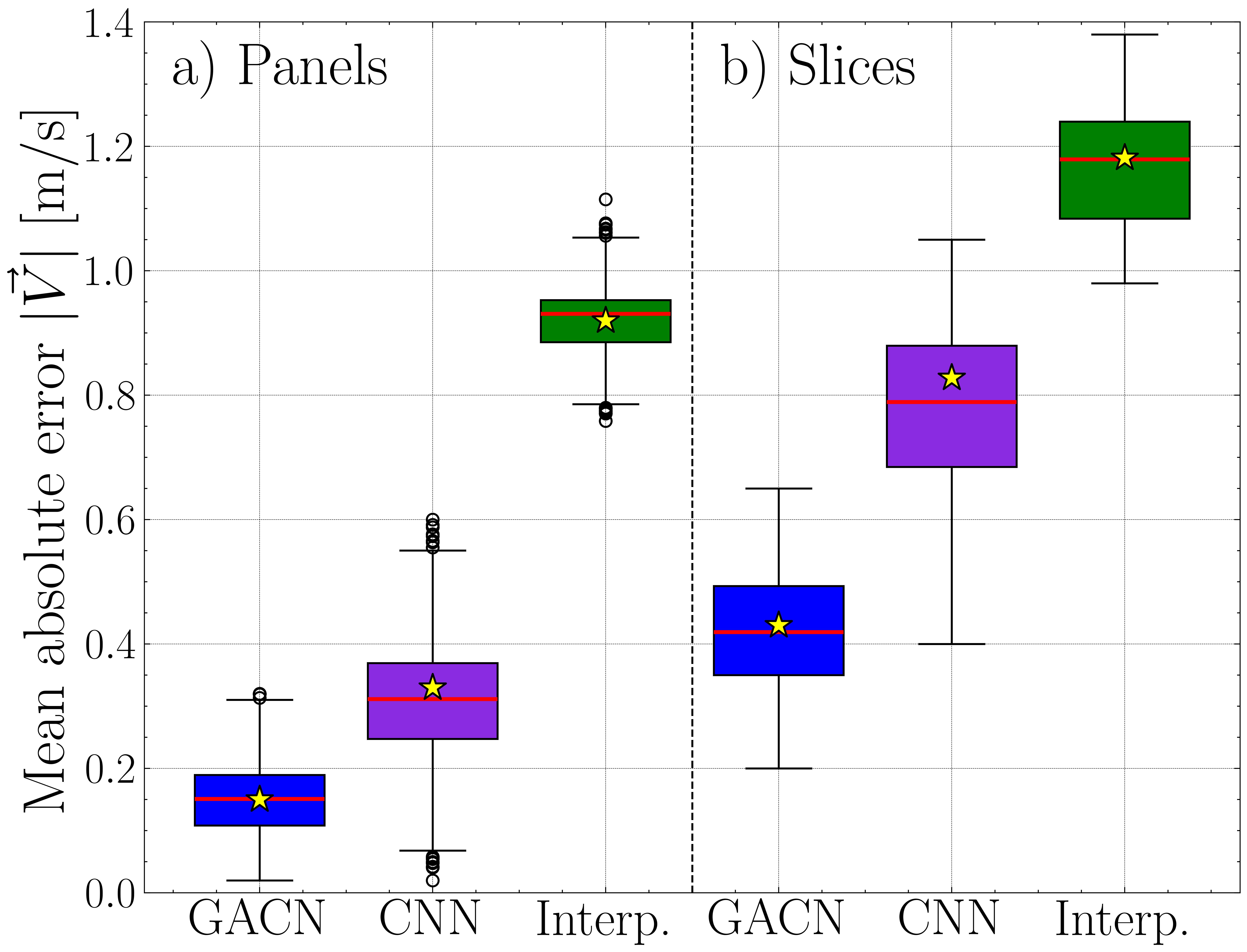}
\caption{\footnotesize Distribution of the mean absolute prediction error of the GACN model (blue), the CNN (purple) and cubic interpolation (green) for a) the DNS panel and b) the DNS slice test sets. The red lines show the median values, while the yellow star represents the mean value.}
\label{fig:dns_mae_box}
\end{figure}

Figure~\ref{fig:dns_mae_box} illustrates the prediction error metrics averaged over the entire DNS test sets, which comprise 2237 test panels and 840 slices. For the panel test set, the GACN achieves an average MAE of \SI[per-mode=symbol]{0.15}{\meter\per\second}, which is significantly lower than the \SI[per-mode=symbol]{0.32}{\meter\per\second} and \SI[per-mode=symbol]{0.92}{\meter\per\second} observed for the CNN and cubic interpolation, respectively. In addition, the GACN exhibits a narrower MAE distribution compared to the CNN, highlighting its effectiveness in minimizing prediction errors and improving the overall reliability of the flow reconstruction.
For the slice test set, which is the larger domain, the performance difference increases further. The GACN maintains its superior performance with an MAE of \SI[per-mode=symbol]{0.42}{\meter\per\second}, while the CNN and interpolation methods yield an MAE of \SI[per-mode=symbol]{0.78}{\meter\per\second} and \SI[per-mode=symbol]{1.18}{\meter\per\second}, respectively. This significant difference in predictive accuracy at larger domains could be partly due to the graph structure of the GACN model. By including the distances between nodes as a training feature, the GACN gains a sense of scale, potentially facilitating better generalization to larger domain sizes.

\subsection{Reconstruction of PIV data\label{subsec:piv_results}}

\begin{figure*}[h!]
\centering
\includegraphics[width=0.6\linewidth]{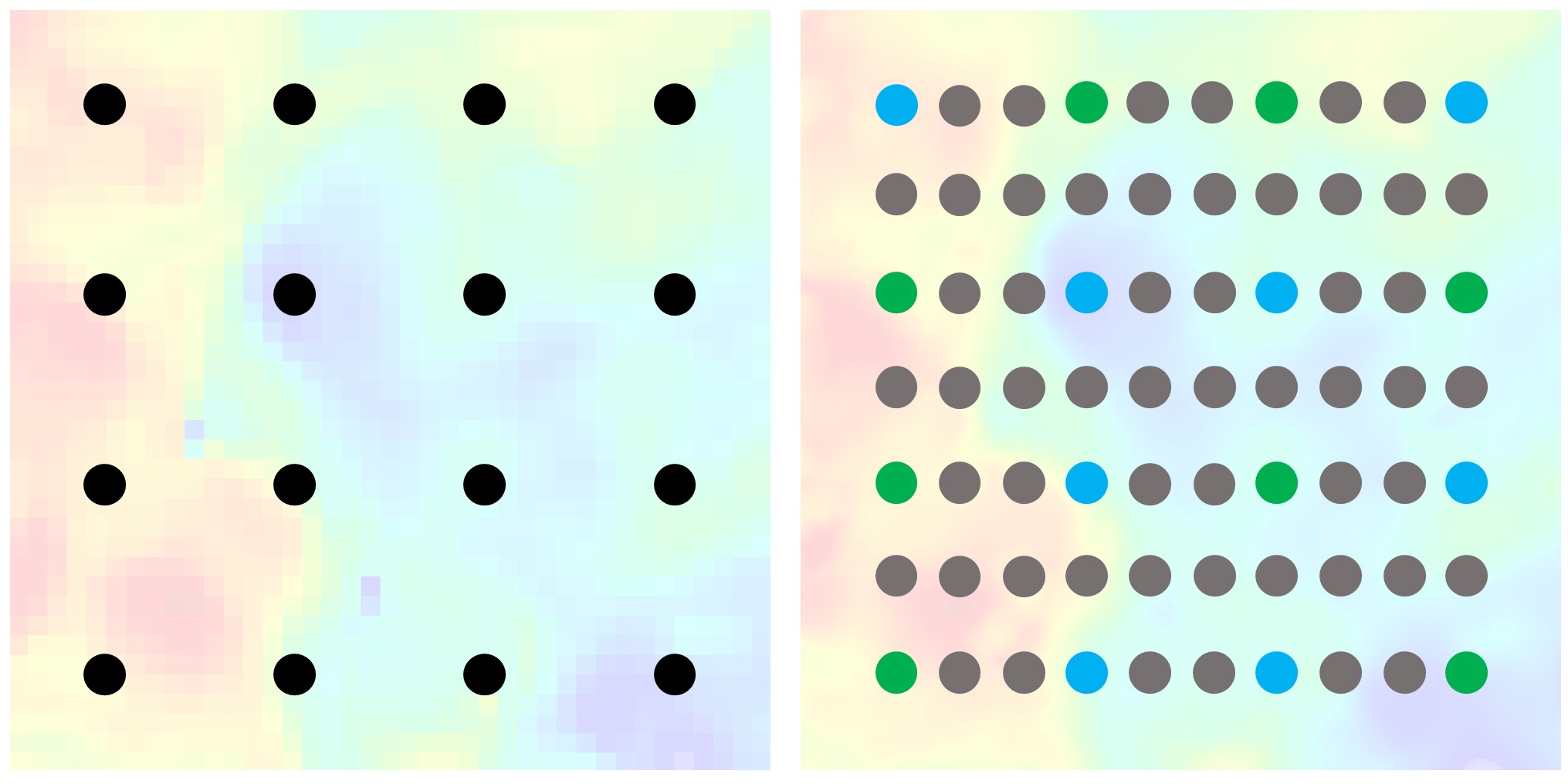}
\caption{\footnotesize Illustration of the data preparation process for the PIV super-resolution-like prediction. Left: original PIV data with black nodes representing the measurement locations. Right: augmented data for reconstruction, where blue points represent input data, green nodes are reserved for error testing and grey nodes are assigned zero velocity.}
\label{fig:piv_sr}
\end{figure*}

The reconstruction capabilities were also tested using the PIV test set without fine tuning or retraining the network, using both $20\times20$~\SI{}{\milli\meter\squared} panels and entire $xz$ slices as input. The PIV data has a significantly lower spatial resolution than the DNS with 1521 grid points on each panel and 16,641 on each $xz$ slice. 
To evaluate the reconstruction accuracy and error of the networks, we randomly split the PIV data into two equal size subsets. One subset was used to generate the input (illustrated by the blue nodes in Fig.\ref{fig:piv_sr}, right.) 
The other half of the points was reserved for error testing (green nodes in Fig.\ref{fig:piv_sr}, right). In order to provide the networks an input with a size closer to what was seen during training, evenly spaced nodes were introduced between the PIV points (grey nodes in Fig.~\ref{fig:piv_sr}, right) increasing the total number of nodes to 76,000 for panels and 832,000 for slices. The features at the intermediate nodes were initialized with zero velocity values. After the addition of these extra nodes, the amount of valid information (non-zero velocity data) in the input was reduced to approximately 1\% of the total nodes to match the sparsity level used in the DNS data set and during network training.

\begin{figure*}[h!]
\centering
\includegraphics[width=\linewidth]{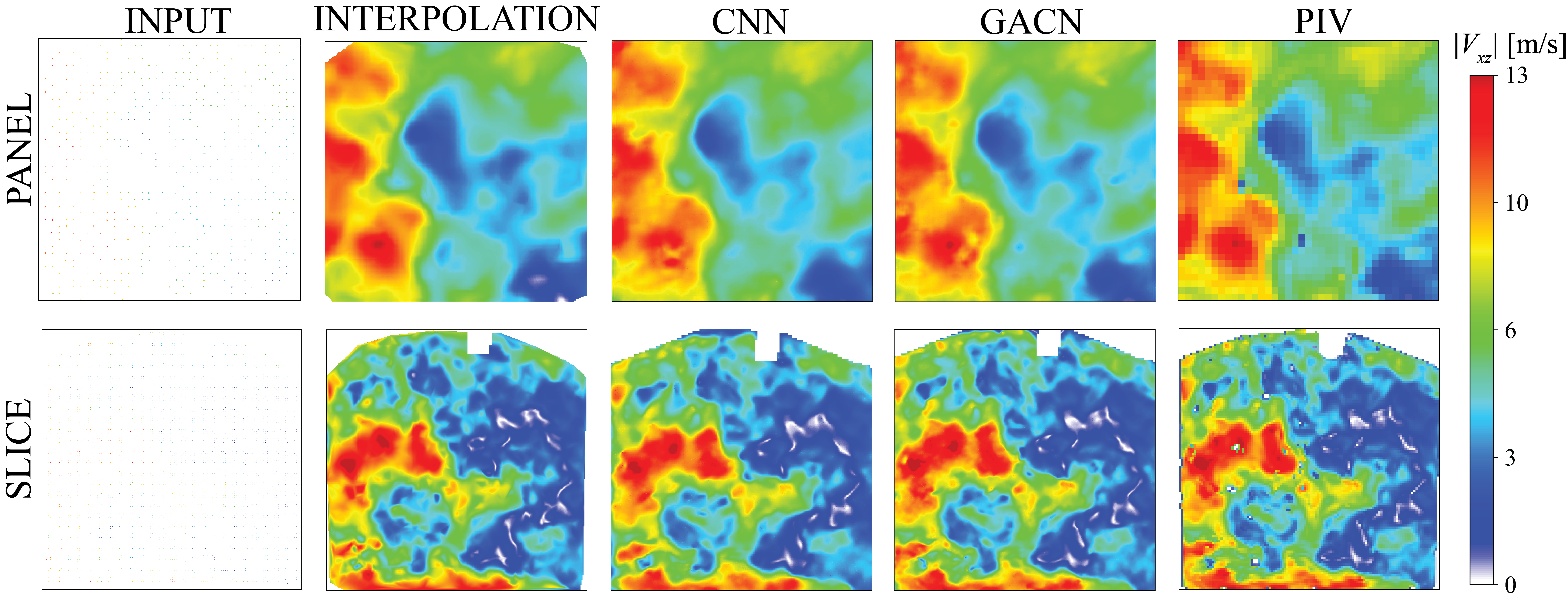}
\caption{\footnotesize Results on one panel at -85 CAD (top row) and one $xz$ slice at -90 CAD (bottom row) from the PIV test set. From left to right: input with 1\% valid information, cubic interpolation baseline prediction, CNN model prediction, GACN model prediction and the low-resolution PIV ground truth.}
\label{fig:piv_panel_slice}
\end{figure*}

To evaluate the reconstruction accuracy of the network for this super-resolution-like task, we compared the predictions only for half of the PIV measurement points that were not used as input (green nodes in Fig.\ref{fig:piv_sr}, right). Figure~\ref{fig:piv_panel_slice} illustrates the results for one of the measured panel and slice. Both networks are able to reconstruct the flow with high accuracy, considering that more than 99\% of the input domain contains zero values. The GACN model shows superior performance in capturing smaller flow features, resulting in a more accurate overall prediction. The computed MAE and RMSE values for the panel and slice, shown in Table~\ref{tab:piv_mae_rmse}, further confirm the improved predictive capabilities of the GACN model. It is important to note that this super-resolution-like task presents an interesting interpretative challenge. 
The ability of the network to generate realistic flow structures in these regions suggests a form of physics-informed interpolation that may provide insights beyond the original measurement resolution. However, this also emphasises the need for careful interpretation of the results, especially in regions for which no PIV data is available.  
Future work could investigate methods to validate these interpolated regions using higher resolution PIV measurements.

\begin{table}[h]
\caption{Prediction errors for the PIV slice and panel shwon in Fig.~\ref{fig:piv_panel_slice}}
\label{tab:piv_mae_rmse}
\centering
\begin{tabular}{lccc}
\hline
Models & MAE [\SI[per-mode=symbol]{}{\meter\per\second}] & RMSE [\SI[per-mode=symbol]{}{\meter\per\second}] \\
\hline
GACN & 0.241 & 0.448 \\
CNN & 0.343 & 0.614 \\
Cubic interpolation & 0.712 & 1.125 \\
\hline
\end{tabular}
\end{table}

\begin{figure}[h!]
\centering
\includegraphics[width=0.6\linewidth]{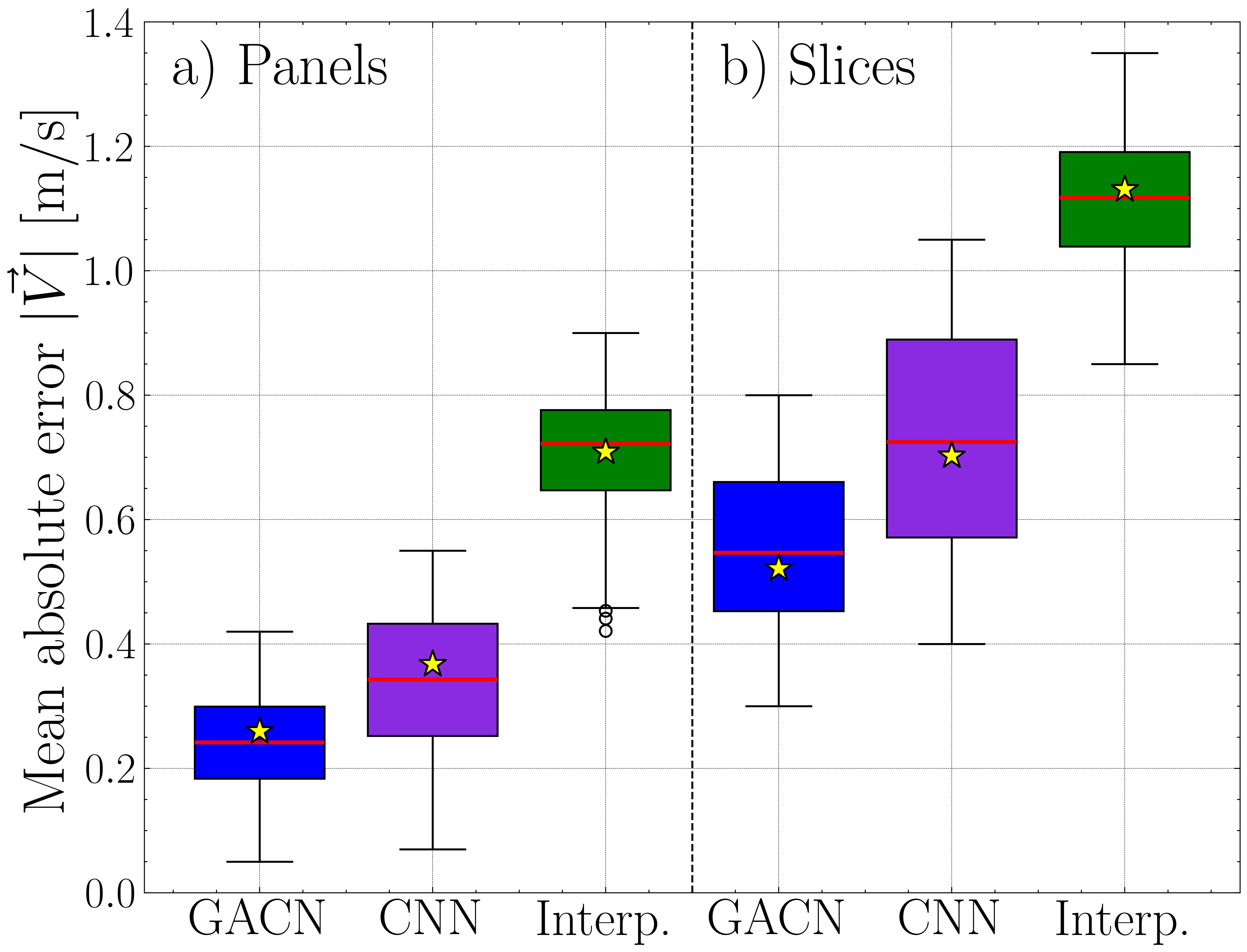}
\caption{\footnotesize Distribution of the mean absolute prediction error of the GACN model (blue), the CNN (purple) and cubic interpolation (green) for a) the PIV panel and b) the PIV slice test sets. The red lines show the median values, while the yellow star represents the mean value.} 
\label{fig:piv_box_mae}
\end{figure}

Figure~\ref{fig:piv_box_mae} shows the error metrics for the entire PIV data set, which includes 612 panels and 204 $xz$ slices. The GACN model outperforms the other methods with lower mean values and narrower error distributions. For the panel predictions, the GACN achieves a MAE of \SI[per-mode=symbol]{0.24}{\meter\per\second}, compared to \SI[per-mode=symbol]{0.34}{\meter\per\second} for the CNN and \SI[per-mode=symbol]{0.71}{\meter\per\second} for the cubic interpolation. Similarly, for the $xz$ slices predictions with the GACN, the CNN and the interpolation methods, the MAE are \SI[per-mode=symbol]{0.55}{\meter\per\second}, \SI[per-mode=symbol]{0.72}{\meter\per\second} and \SI[per-mode=symbol]{1.14}{\meter\per\second}, respectively.
The difference in performance between the ML-based approaches is less pronounced than in the DNS data set, possibly due to the structured nature of both the PIV data and the synthetically added points. This structured arrangement may reduce the advantage of the graph structure of the GACN, which typically excels with unstructured data.
Interestingly, the interpolation errors for the PIV data set are slightly lower than for the DNS data. This improvement is probably due to the lower resolution of the PIV data, which inherently captures fewer small-scale turbulent structures compared to the high-resolution DNS data. When interpolation is applied to the PIV data, the absence of these small-scale structures in the original measurements leads to a more accurate reconstruction. 

\section{Conclusions\label{sec:conclusions}}

The paper presents a Graph Attention Convolutional Network (GACN) for flow reconstruction from very sparse data. The model employs a preprocessing step based on a Feature Propagation (FP) algorithm that effectively handles sparse inputs by leveraging information from neighboring nodes, improving the initialization of missing features, and ensuring physical consistency. Additionally, we introduce a binary indicator (BI) as an extra feature, serving as a validity mask to differentiate between original and propagated data points. This combination enables more effective learning from sparse inputs.  

The network is trained and tested on one of the largest DNS data sets for Internal Combustion Engines (ICE), showcasing its ability to handle unstructured data and perform well at different resolutions and domain sizes. A key feature of the model is its capacity to effectively process time-varying engine data, where the computational domain changes due to piston motion. This results in variable input sizes throughout the engine cycle, which the GACN can seamlessly accommodate without requiring data padding or interpolation. Notably, the model exhibits robust generalization capabilities, performing effectively even when tested with experimental PIV data that were not used during training. 

The comparative analysis shows that the GACN consistently outperforms both a conventional CNN and cubic interpolation methods. The GACN achieves lower reconstruction errors on both the DNS and PIV test sets, with the additional major advantage of not requiring data interpolation onto a structured grid, as is necessary for CNN. This ability to process unstructured data directly makes the GACN particularly suitable for complex geometries and time-varying domains, which are characteristic of ICE. In addition, the GACN model effectively reconstructs flow fields not only from small panels used in training, but also from entire slices with domains up to 14 times larger than those seen during the training phase. A key factor contributing to this generalization ability is likely the inclusion of distance features between nodes in the graph structure, giving the network a sense of scale and allowing it to adapt to larger domains by understanding the spatial relationships between data points, regardless of the overall size of the domain. The difference between GACN and other methods becomes more pronounced in these larger domains, highlighting its superior ability to capture and reconstruct complex flow patterns across different spatial scales.

While this work focuses on flow reconstruction in ICE, the developed GACN model can potentially have broader applicability for various engineering applications where reconstruction from sparse or insufficiently resolved data is required, especially for unstructured domains. As a next step, we will investigate the predictive and modeling capabilities of the GACN for near-wall phenomena~\citep{dupuy2023}. An extension of the approach to full 3D flow field reconstruction is also of interest at the expense of a significant increase in the computational cost of model training. The application to reactive flows, using a newly developed version of the reactive flow solver \citep{Kinetix, nekCRF}, would be particularly interesting. This is challenging for various reasons, as in fast chemistry the source terms depend on the smallest scales, which means that these must be predicted correctly for such multi-scale fields.
 
\section*{Acknowledgments}
This project received funding from the European Union’s Horizon 2020 research and innovation program under the Center of Excellence in Combustion (CoEC) project, grant agreement No 952181. The authors gratefully acknowledge the Gauss Centre for Supercomputing e.V. for funding this project by providing computing time on the GCS Supercomputer JUWELS at Julich Supercomputing Centre (JSC). MS and BB acknowledge support by Deutsche Forschungsgemeinschaft through SFB-Transregio 150 Project Number 237267381-TRR150.

\section*{Funding} 
This work received funding from the European Union’s Horizon 2020 research and innovation program under the Center of Excellence in Combustion (CoEC) project, grant agreement No 952181 as well as the Forschungsvereinigung Verbrennungskraftmaschinen (FVV, project no. 6015252) and the Swiss Federal Office of Energy (BfE, contract no. SI/502670-01) under the project "WHTforH2".

\section*{Declaration of competing interest}
The authors declare that they have no known competing financial interests or personal relationships that could influence the work reported in this paper.

\section*{CRediT authorship contribution statement} \textbf{Bogdan A. Danciu}: Conceptualization, Methodology, Software, Validation, Formal analysis, Data curation, Visualization, Writing - original draft, Writing - review and editing. \textbf{Vito A. Pagone}: Conceptualization, Methodology, Software, Validation, Data curation, Writing - review and editing. \textbf{Benjamin Böhm}: Validation, Experimental data curation, Writing - review and editing. \textbf{Marius Schmidt}: Validation, Experimental data curation, Writing - review and editing. \textbf{Christos E. Frouzakis}: Conceptualization, Validation, Funding acquisition, Resources, Project administration, Writing - review and editing. 

\section*{Data Availability} 
The data are available from the authors upon reasonable request.

\bibliographystyle{elsarticle-harv}
\bibliography{refs}

\end{document}